\documentclass[letterpaper]{article}

\usepackage{aaai}
\usepackage{times} 
\usepackage{helvet} 
\usepackage{courier} 

\usepackage{color}
\usepackage{amsmath}
\usepackage{amssymb}
\usepackage{amsthm}
\usepackage[ruled]{algorithm}
\usepackage[noend]{algpseudocode}
\usepackage{graphicx}
\usepackage{colortbl}
\usepackage{arydshln}

\usepackage{tikz}
\usetikzlibrary{positioning}
\usetikzlibrary{shapes.geometric}
\usetikzlibrary{shapes.symbols}
\usetikzlibrary{shadows}
\usetikzlibrary{arrows}

\title{Maximum a Posteriori Estimation by Search in Probabilistic Programs}
\author {David Tolpin, Frank Wood\\University of Oxford\\ \{dtolpin,fwood\}@robots.ox.ac.uk}
\begin{document}

\maketitle

\begin{abstract}
  We introduce an approximate search algorithm for fast {\it maximum a
    posteriori} probability estimation in probabilistic
  programs, which we call Bayesian ascent Monte Carlo (BaMC).
  Probabilistic programs represent probabilistic models with varying
  number of mutually dependent finite, countable, and continuous
  random variables. BaMC is an anytime MAP search algorithm applicable
  to any combination of random variables and dependencies.  We compare
  BaMC to other MAP estimation algorithms and show that BaMC is faster
  and more robust on a range of probabilistic models. 
\end{abstract}

\section{Introduction}

Many Artificial Intelligence problems, such as approximate planning in
MDP and POMDP, probabilistic abductive
reasoning~\cite{R11}, or utility-based recommendation~\cite{SG09}, can
be formulated as MAP estimation problems. The framework of
probabilistic inference~\cite{P88} proposes solutions to a wide range
of Artificial Intelligence problems by representing them as
\textit{probabilistic models}. Efficient domain-independent algorithms
are available for several classes of representations, in particular
for graphical models~\cite{L96}, where inference can be performed
either exactly and approximately.  However, graphical models typically
require that the full graph of the model to be represented explicitly,
and are not powerful enough for problems where the state space is
exponential in the problem size, such as the generative models common
in planning~\cite{SKM14}.

Probabilistic programs~\cite{GMR+08,WVM14} can represent
arbitrary probabilistic models. In addition to expressive power,
probabilistic programming separates modeling and inference, allowing
the problem to be specified in a simple language which does not assume
any particular inference technique. Recent success in PMCMC methods
enables efficient sampling from posterior distributions with few
restrictions on the structure of the models~\cite{WVM14,PWD+14}.

However, an efficient sampling scheme for finding a MAP estimate would
be different from the scheme for inferring the posterior distribution:
only a single instantiation of model's variables, rather than their
joint distribution, must be found.  This difference reminds of the
difference between simple and cumulative reward optimization in many
settings, for example, in Multi-armed bandits~\cite{SBM11}: when all
samples contribute to the total reward, the algorithms are said to
optimize the {\it cumulative reward}, which is the classical
Multi-armed bandit settings.  Alternatively, when only the quality of
the final choice matters, the algorithms are said to optimize the {\it
  simple reward}. This setting is often called a {\it search problem}.
Previous research demonstrated that different sampling schemes work
better for either cumulative or simple reward, and algorithms which
are optimal in one setting can be suboptimal in the
other~\cite{HRT+12}.

In this paper, we introduce a sampling-based search algorithm for fast
MAP estimation in probabilistic programs, Bayesian ascent Monte Carlo
(BaMC), which can be used with any combination of finite, countable
and continuous random variables and any dependency structure. We
empirically compare BaMC to other feasible MAP estimation
algorithms, showing that BaMC is faster and more robust.

\section{Preliminaries}

\subsection{Probabilistic Programming}

Probabilistic programs are regular programs extended by two
constructs~\cite{GHNR14}: a) the ability to draw random values from probability
distributions, and  b) the ability to condition values computed in the
programs on probability distributions.  A probabilistic program
implicitly defines a probability distribution over program state. Formally, a
probabilistic program is a \emph{stateful deterministic computation}
$\mathcal{P}$ with the following properties:

\begin{itemize}
    \item Initially, $\mathcal{P}$ expects no arguments.
    \item On every invocation, $\mathcal{P}$ returns either a
        distribution $F$, a distribution and a value $(G, y)$, a
        value $z$, or $\bot$.
    \item Upon returning $F$, $\mathcal{P}$ expects a value $x$
        drawn from $F$ as the argument to continue.
    \item Upon returning $(G, y)$ or $z$, $\mathcal{P}$ is
        invoked again without arguments.
    \item Upon returning $\bot$, $\mathcal{P}$ terminates.
\end{itemize}

A program is run by calling $\mathcal{P}$ repeatedly until
termination. Every run of the program implicitly produces a
sequence of pairs $(F_i, x_i)$ of distributions and values drawn
from them. We call this sequence a \textit{trace} and denote it
by $\pmb{x}$.  Program output is deterministic given
the trace.

By definition, the probability of a trace is proportional
to the product of the probability of all random choices
$\pmb{x}$ and the likelihood of all observations $\pmb{y}$:
\begin{equation}
    p_{\mathcal{P}}(\pmb{x}|\pmb{y}) \propto \prod_{i=1}^{\left|\pmb{x}\right|}
    p_{F_i}(x_i) \prod_{j=1}^{\left|\pmb{y}\right|}p_{G_j}(y_{j})
  \label{eqn:p-trace}
\end{equation}
The objective of inference in probabilistic program
$\mathcal{P}$ is to discover the distribution
of program output. 
\footnote{Note that this conceptualization of a probabilistic program corresponds, for example, to the approach in~\cite{GS15}.}

Several implementations of general probabilistic programming
languages are available~\cite{GMR+08,WVM14}.  Inference is
usually performed using Monte Carlo sampling algorithms for
probabilistic programs~\cite{WSG11,WVM14,PWD+14}.  While some
algorithms are better suited for certain problem types, most can
be used with any valid probabilistic program.

\subsection{Maximum a Posteriori Probability Inference}

Maximum \textit{a posteriori} probability (MAP) inference is the
problem of finding an assignment to the variables of a probabilistic
model that maximizes their joint posterior probability~\cite{M12}.
Sometimes, a more general problem of marginal MAP inference
estimation is solved, when the distribution is marginalized over
some of the variables~\cite{DGR02,MC12}. In this paper we consider
the simpler setting of MAP estimation, where assignment for all
variables is sought, however the proposed algorithms can be extended to
marginal MAP inference.

For certain graphical models the MAP assignment can be found
exactly~\cite{PD03,SDY07}.  However, in most advanced cases, e.g.
models expressed by probabilistic programs, MAP
inference is intractable, and approximate algorithms such as
Stochastic Expectation-Maximization~\cite{WT90} or Simulated
Annealing~\cite{AD00} are used.

Simulated Annealing (SA) for MAP inference constitutes a universal
approach which is based on Monte Carlo sampling. Simulated
Annealing is a non-homogeneous version of Metropolis-Hastings
algorithm where the acceptance probability is gradually changed
in analogy with the physical process of annealing~\cite{KGV83}.
Convergence of Simulated Annealing algorithms depends on the
properties of the \textit{annealing schedule} --- the rate with
which the acceptance probability changes in the course of the
algorithm~\cite{LM86}. When the rate is too low, the SA algorithm
may take too many iterations to find the global maximum. When
the rate is too high, the algorithm may fail to find the global
maximum at all and get stuck in a local maximum instead.
Tuning the annealing schedule is necessary to achieve
reasonable performance with SA, and the best schedule depends on
both the problem domain and model parameters.

\section{Bayesian Ascent Monte Carlo}

We introduce here an approximate search algorithm for fast MAP
estimation in probabilistic programs, Bayesian ascent Monte Carlo
(BaMC). The algorithm draws inspiration from Monte Carlo Tree
Search~\cite{KS06}. Unlike Simulated Annealing, BaMC uses the
information about the probability of every sample to propose
assignments in future samples, a kind of
\textit{adaptive proposal} in Monte Carlo inference. BaMC
differs from known realizations of MCTS  in a number of ways.

\begin{itemize}
    \item The first difference between BaMC and MCTS as commonly
implemented in online planning or game playing follows from the
nature of inference in probabilistic models. In online planning
and games, the search is performed with the root of the search
corresponding to the current state of the agent. After a certain
number of iterations, MCTS commits to an action, and restarts
the search for the action to take in the next state. In
probabilistic program inference assignment to all variables must
be determined simultaneously, hence the sampling is always
performed for all variables in the model. 

\item Additionally, probabilistic programs often involve a combination
of finite, infinite countable, and infinite continuous random
variables. Variants of MCTS for continuous variables were
developed~\cite{CHS+11}, however mixing variables of different
types in the same search is still an open problem. BaMC uses
\textit{open randomized probability matching}, also introduced
here, to handle all variables in a unified way independently of variable 
type.

\item Finally, BaMC is an any-time algorithm. Since BaMC searches for
an estimate of the maximum of the posterior probability, every
sample with a greater posterior probability than that of all
previous samples is an improved estimate. BaMC outputs all such
samples. As sampling goes on, the quality of solution improves,
however any currently available solution is a MAP estimate of
the model with increasing quality.
\end{itemize}

BaMC (Algorithm~\ref{alg:bamc}) maintains beliefs about
probability distribution of \textit{log weight} (the logarithm of unnormalized 
probability defined by Equation~\ref{eqn:p-trace}) of the trace
for each value of each random variable in the probabilistic
program.  At every iteration (Algorithm~\ref{alg:bamc}) the
algorithm runs the probabilistic program
(lines~\ref{alg:bamc-P-start}--\ref{alg:bamc-P-end}) and
computes the log weight of the trace.  If the log weight of the
trace is greater than the previous maximum log weight, the
maximum log weight is updated, and the trace is output as a new
MAP estimate~(lines~\ref{alg:bamc-mape-start}--\ref{alg:bamc-mape-end}).
Finally, the beliefs are updated from the log weight of the
sample~(lines~\ref{alg:bamc-update-start}--\ref{alg:bamc-update-end}).

\begin{algorithm}[t]
\caption{Monte Carlo search for MAP assignment.}
\label{alg:bamc}
\begin{algorithmic}[1]
  \State max-log-weight $\gets -\infty$
  \Loop
    \State trace $\gets$ (), log-weight $\gets$ 0
	\State result $\gets$ \Call{$\mathcal{P}$}{{}} \hspace{4em}{\color{blue!35!white} /* {\em probabilistic program} */} \label{alg:bamc-P-start}
    \Loop         
      \If {result \textbf{is} $F_i$}
         \State $x_i$ $\gets$ \Call{SelectValue}{$i$, $F_i$} \label{alg:bamc-select}
         \State log-weight $\gets$ log-weight + $\log p_{F_i}(x_i)$
         \State log-weight$_i$ $\gets$ log-weight
         \State \Call{Push}{trace,($F_i$, $x_i$)}
         \State result $\gets$ \Call{$\mathcal{P}$}{$x_i$}
      \ElsIf {result \textbf{is} $(G_j, y_j)$}
        \State log-weight $\gets$ log-weight + $\log p_{G_j}(y_j)$
        \State result $\gets$ \Call{$\mathcal{P}$}{{}}
      \ElsIf {result \textbf{is} $z_k$}
         \State \Call{Output}{$z_k$}
         \State result $\gets$ \Call{$\mathcal{P}$}{{}}
	  \Else {\bf{ }break}
      \EndIf
    \EndLoop \label{alg:bamc-P-end}
    \If {log-weight $>$ max-log-weight} \label{alg:bamc-mape-start}
        \State \Call{Output}{trace}
        \State max-log-weight $\gets$ log-weight
    \EndIf \label{alg:bamc-mape-end}
	\For {$i$ \textbf{in} $|$trace$|$ \textbf{downto} 1}
	\label{alg:bamc-update-start}
       \State \Call{Update}{i, log-weight - log-weight$_i$}
	   \label{alg:bamc-update-end}
    \EndFor
  \EndLoop
\end{algorithmic}
\end{algorithm}

The ability of BaMC to discover new, improved MAP estimates
depends on the way values are selected for random
variables (line~\ref{alg:bamc-select}). On one hand, new values
should be drawn to explore the domain of the random
variable. On the other hand, values which were tried previously
and resulted in a high-probability trace should be re-selected
sufficiently often to discover high-probability assignments
conditioned on these values.

\subsection{Open Randomized Probability Matching}

Randomized probability matching (RPM), also called Thompson
sampling~\cite{T33}, is used in many contexts where choices are
made based on empirically determined choice rewards.  It is a
selection scheme that maintains beliefs about reward
distributions of every choice, selects a choice with the
probability that the average reward of the choice is the highest
one, and revises beliefs based on observed rewards. Bayesian
belief revision is usually used with randomized probability
matching. Selection can be implemented efficiently by drawing a
single sample from the belief distribution of average belief for
every choice, and selecting the choice with the highest sample
value~\cite{S10,AG12}.

Here we extend randomized probability matching to domains of
infinite or unknown size. We call this generalized version
\textit{open randomized probability matching} (ORPM)
(Algorithm~\ref{alg:orpm}). ORPM is given a choice distribution,
and selects choices from the distribution to maximize the total
reward. ORPM does not know or assume the type of the choice
distribution, but rather handles all distribution types in a unified
way. Like RPM, ORPM maintains beliefs about the rewards of every 
choice. First, ORPM uses RPM to guess the reward distribution of
a randomly drawn choice
(lines~\ref{alg:orpm-guess-start}--\ref{alg:orpm-guess-end}).
Then, ORPM uses RPM again to select a choice based on the
beliefs of each choices, \textit{including a randomly drawn
choice}
(lines~\ref{alg:orpm-select-start}--\ref{alg:orpm-select-end}).
If the selected choice is a randomly drawn choice,
the choice is drawn from the choice distribution
(line~\ref{alg:orpm-draw}) and added to the set of choices.
Finally, an action is executed based on the choice
(line~\ref{alg:orpm-execute}) and the reward belief of the
choice is updated based on the reward updated from the action.

\begin{algorithm}[t]
\caption{Open randomized probability matching.}
\label{alg:orpm}
\begin{algorithmic}[1]
    \State choices $\gets$ ()
    \Loop
	\State {\color{blue!35!white} /* $x_i$ $\gets$ \Call{SelectValue}{$i$, $F_i$} */ }
   	  \If {choices = ()} {{ }best-choice $\gets$ \textit{random-choice}} \label{alg:orpm-select-start}
        \Else
			\State best-reward $\gets -\infty$ \label{alg:orpm-guess-start}
            \State best-belief $\gets \bot$ 
            \For {choice \textbf{in} choices}
                \State reward $\gets$ \Call{Draw}{RewardBelief(choice)}
                \If {reward $\ge$ best-reward}
                   \State best-reward$\gets$reward
				   \State best-belief{}$\gets$MeanRewardBelief(choice)
                \EndIf
				\EndFor \label{alg:orpm-guess-end}
            \State best-reward $\gets$ \Call{Draw}{best-belief}
            \State best-choice $\gets$ \textit{random-choice}
            \For {choice \textbf{in} choices} \label{alg:orpm-rpm-start}
			\State\!reward$\gets$\Call{Draw}{MeanRewardBelief(choice)}
                \If {reward $\ge$ best-reward}
                   \State best-reward $\gets$ reward
                   \State best-choice $\gets$ choice
                \EndIf
			\EndFor \label{alg:orpm-rpm-end}
        \EndIf \label{alg:orpm-select-end}
        \If {best-choice = \textit{random-choice}}
            \State best-choice $\gets$ \Call{Draw}{ChoiceDistribution} \label{alg:orpm-draw}
            \State RewardBelief(best-choice) $\gets$ PriorBelief
            \State choices $\gets$ \Call{Append}{choices,best-choice} \label{alg:orpn-append-choice}
			\EndIf \label{alg:orpm-select-end}
			\State {\color{blue!35!white} /* result $\gets\mathcal{P}()$ */ }
        \State reward $\gets$ \Call{Execute}{best-choice} \label{alg:orpm-execute}
		\State {\color{blue!35!white} /* \Call{Update}{i, log-weight - log-weight$_i$} */}
        \State \Call{UpdateRewardBelief}{best-choice,reward} \label{alg:orpm-update}
    \EndLoop
\end{algorithmic}
\end{algorithm}

The final form of Bayesian Ascent Monte Carlo is obtained by
combining Monte Carlo search for MAP assignment
(Algorithm~\ref{alg:bamc}) and open randomized probability
matching (Algorithm~\ref{alg:orpm}). Selecting a value in
line~\ref{alg:bamc-select} corresponds to
lines~\ref{alg:orpm-select-start}--\ref{alg:orpm-select-end} of
Algorithm~\ref{alg:orpm}. Line~\ref{alg:orpm-update} of
Algorithm~\ref{alg:orpm} is performed at every iteration of the
loop in
lines~\ref{alg:bamc-update-start}--\ref{alg:bamc-update-end} of
Algorithm~\ref{alg:bamc}. The reliance on randomized probability
matching allows to implement BaMC efficiently without any
tunable parameters.

\subsection{Belief Maintenance}

In probabilistic models with continuous random choices $\log
p_\mathcal{P}(\pmb{x|y})$ (Equation~\ref{eqn:p-trace}) may be both
positive and negative, and is in general unbounded on either side,
therefore we opted for the normal distribution to represent
beliefs about choice rewards.  Since parameters of reward
distributions vary in wide bounds, and reasonable initial
estimates are hard to guess, we used an uninformative prior
belief, which is in practice equivalent to maintaining sample
mean and variance estimates for each choice, and using the
estimates as the parameters of the belief distribution. Let us
denote by $r_{ij}$  a random variable corresponding to the
reward attributed to random choice $j$ at selection point $i$.
Then the reward belief distribution is
\begin{equation}
    r_{ij} \sim Bel(r_{ij})\triangleq\mathcal{N}(\mathrm{E}(r_{ij}), \mathrm{Var}(r_{ij})),
    \label{eqn:reward-belief}
\end{equation}
where $\mathrm{E}(r_{ij})$ and $\mathrm{Var}(r_{ij})$ are the sample
mean and variance of the reward, correspondingly. 

In the same uninformative prior setting, the \textit{mean
reward} belief $r^{mean}_{ij}$, used by randomized probability
matching, is
\begin{equation}
    r^{mean}_{ij} \sim Bel(r^{mean}_{ij})
    \triangleq\mathcal{N}\left(\mathrm{E}(r_{ij}), \frac {\mathrm{Var}(r_{ij})} n\right)
    \label{eqn:mean-reward-belief}
\end{equation}
where $n$ is the sample size of $r_{ij}$~\cite{GCS+03}.

These beliefs can be computed efficiently, and provide sufficient
information to guide MAP search. Informative priors on
reward distributions can be imposed to improve convergence when
available, such as in the approach described in~\cite{BBW+13}.

\section{Empirical Evaluation}

\begin{figure}[t]
	\centering
    \includegraphics[scale=0.43]{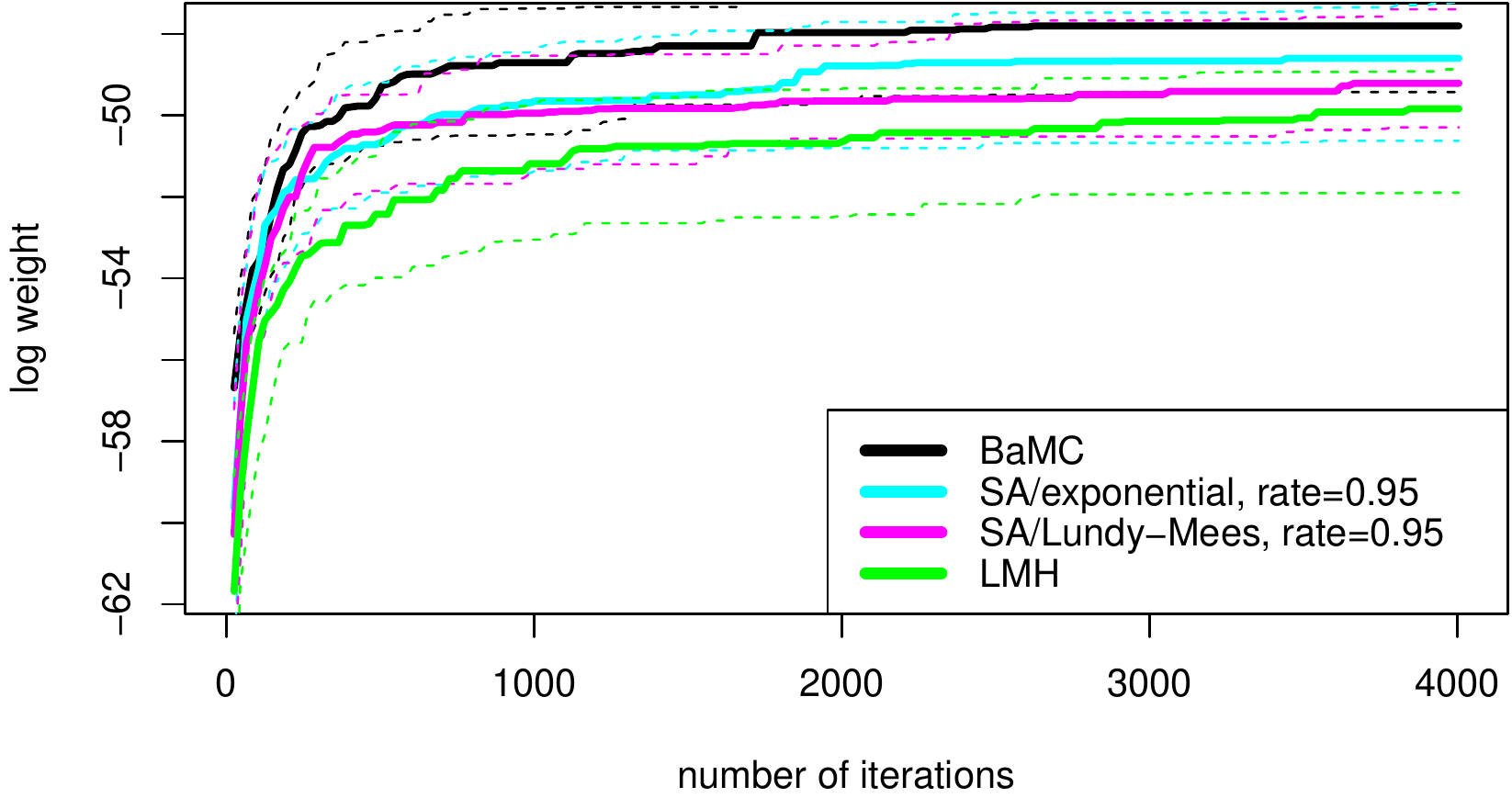}
    \caption{HMM with 16 observed states and unknown transition probabilities.}
    \label{fig:hmm}
\end{figure}

\begin{figure}[t]
    \centering
    \includegraphics[scale=0.43]{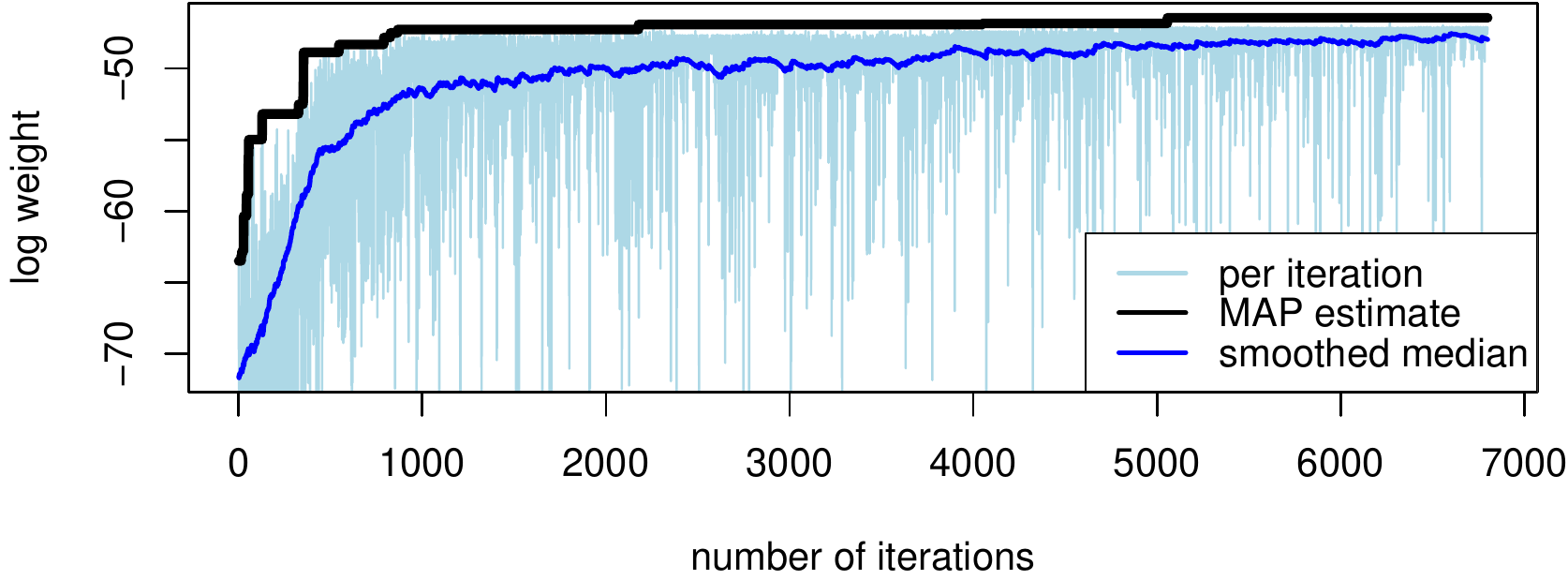}
    \caption{A single run of BaMC on HMM.}
    \label{fig:hmm-bamc}
\end{figure}

We present here empirical results for MAP estimation on two
problems, Hidden Markov Model with 16 observable states and
unknown transition probabilities
(Figures~\ref{fig:hmm},~\ref{fig:hmm-bamc}) and Probabilistic
Infinite Deterministic Automata~\cite{PBW10}, applied to the first chapter of
``Alice's Adventures in Wonderland'' as the training data. Both
represent, for purposes of MAP estimation, probabilistic models
of reasonable size, with a mix of discrete and continuous random
variables.

\begin{figure}[t]
    \centering
    \includegraphics[scale=0.43]{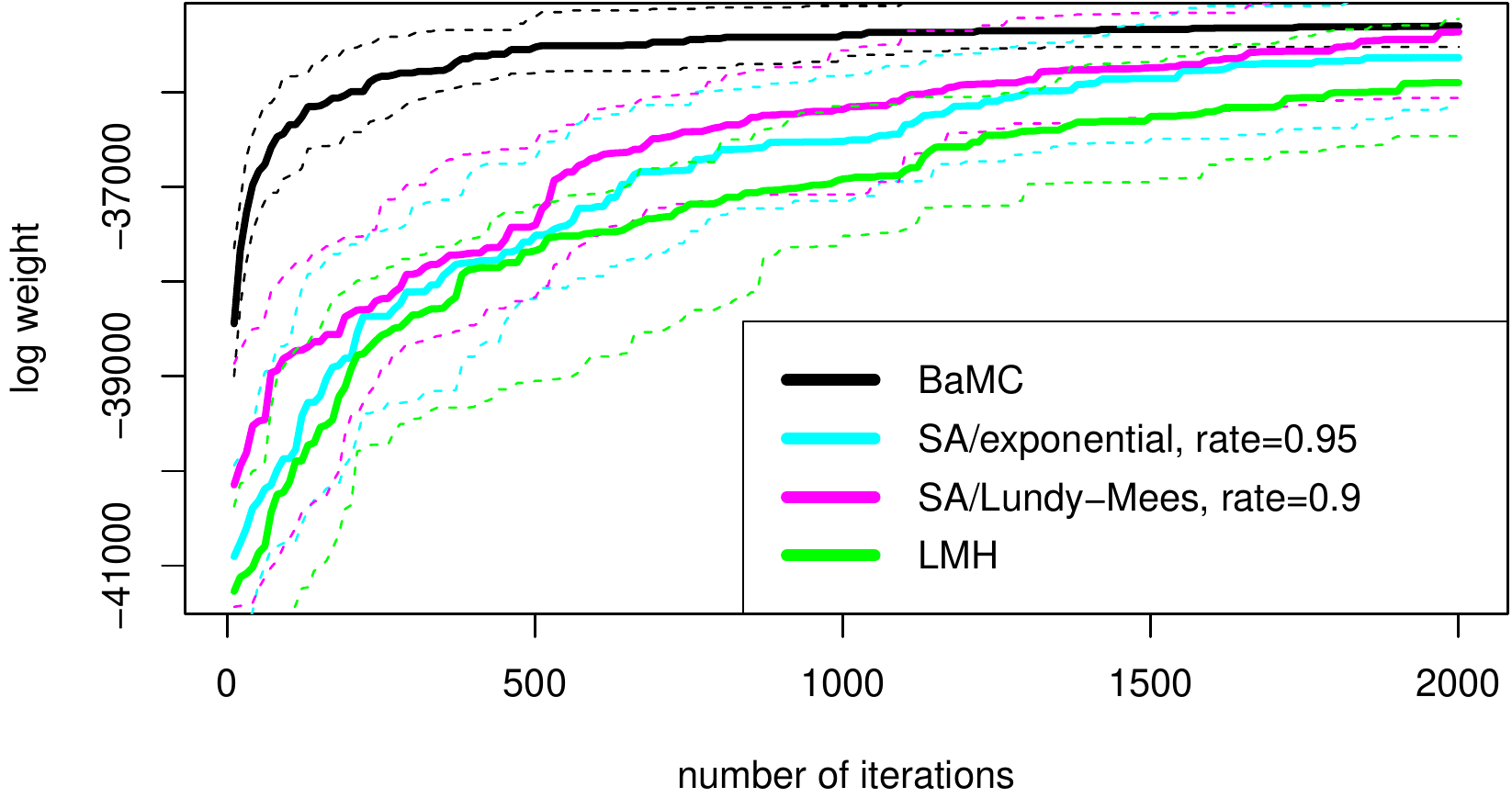}
    \caption{Probabilistic Deterministic Infinite Automata on the
    first chapter of Alice's Adventures in Wonderland.}
    \label{fig:pdia}
\end{figure}

For both problems, we compared BaMC, Simulated Annealing with
exponential schedule and the schedule used in~\cite{LM86} which
we customarily call Lundy-Mees schedule, as well a lightweight
implementation of Metropolis-Hastings~\cite{WSG11} adapted for
MAP search, as a base line for the comparison. In 
Figures~\ref{fig:hmm},~\ref{fig:pdia} the solid lines correspond to
the medians, and the dashed lines to the 25\% and 75\% quantiles of
MAP estimates produced by each  of the algorithms over 50 runs
for 4000 iterations. For each annealing schedule of SA we kept
only lines corresponding to the empirically best annealing rate
(choosing the rate out of the list 0.8, 0.85, 0.9, 0.95). In
both case studies, BaMC consistently outperformed other
algorithms, finding high probability MAP estimates faster and
with less variability  between runs.

In addition, Figure~\ref{fig:hmm-bamc} visualizes a single run
of BaMC on HMM. The solid black line shows the produced MAP
estimates, the light-blue lines are weights of individual
samples, and the bright blue line is the smoothed median of
individual sample weights. One can see that the smoothed median
approaches the MAP estimate as the sampling goes on, reflecting
the fact the BaMC samples gradually converge to a small set of
high probability assignments.

\section{Discussion}

In this paper, we introduced BaMC, a search algorithm for fast MAP
estimation. The algorithm is based on MCTS but differs in a
number of important ways. In particular, the algorithm can
search for MAP in models with any combination of variable types,
and does not have any parameters that has to be tuned for a
particular problem domain or configuration.  
As a part of BaMC we introduced open randomized probability
matching, an extension of randomized probability matching to
arbitrary variable types.

BaMC is simple and straightforward to implement both for MAP
estimation in probabilistic programs, and for stochastic
optimization in general. Empirical evaluation showed
that BaMC outperforms Simulated Annealing despite the ability to
tune the annealing schedule and rate in the latter.  BaMC coped
well with cases of both finite and infinite continuous variables
present in the same problem.

Full analysis of algorithm properties and convergence is still a
subject of ongoing work. Conditions under which BaMC converges,
as well as the convergence rate, in particular in the continuous
case, still need to be established, and may shed light on the
boundaries of applicability of the algorithm. On the other hand,
techniques used in BaMC, in particular open randomized
probability matching, span beyond MAP estimation and stochastic
optimization, and may constitute a base for a more powerful
search approach in continuous and mixed spaces in general.

\section{Acknowledgments}
 
This work is supported under DARPA PPAML through the U.S. AFRL under
Cooperative Agreement number FA8750-14-2-0004.

\bibliographystyle{aaai}
\bibliography{refs}

\end{document}